%% file: sn-article.tex
\documentclass[sn-mathphys-num]{sn-jnl}

\usepackage{graphicx}%
\usepackage{multirow}%
\usepackage{amsmath,amssymb,amsfonts}%
\usepackage{amsthm}%
\usepackage{mathrsfs}%
\usepackage[title]{appendix}%
\usepackage{xcolor}%
\usepackage{textcomp}%
\usepackage{manyfoot}%
\usepackage{booktabs}%
\usepackage{algorithm}%
\usepackage{algorithmicx}%
\usepackage{algpseudocode}%
\usepackage{listings}%

\usepackage{siunitx} 

\theoremstyle{thmstyleone}%
\theoremstyle{thmstyletwo}%

\theoremstyle{thmstylethree}%

\begin{document}

\title[Article Title]{SparseFocus: Learning-based One-shot Autofocus for Microscopy with Sparse Content}


\author*[1]{\fnm{Yongping} \sur{Zhai}}\email{zhaiyongping08@nudt.edu.cn}
\equalcont{These authors contributed equally to this work.}

\author[1]{\fnm{Xiaoxi} \sur{Fu}}
\equalcont{These authors contributed equally to this work.}

\author[1]{\fnm{Qiang} \sur{Su}}
\equalcont{These authors contributed equally to this work.}

\author[1]{\fnm{Jia} \sur{Hu}}

\author[1]{\fnm{Yake} \sur{Zhang}}

\author[1]{\fnm{Yunfeng} \sur{Zhou}}

\author[1]{\fnm{Chaofan} \sur{Zhang}}

\author[1]{\fnm{Xiao} \sur{Li}}

\author[1]{\fnm{Wenxin} \sur{Wang}}

\author*[2]{\fnm{Dongdong} \sur{Wu}}\email{wudd01@tmmu.edu.cn}

\author*[3]{\fnm{Shen} \sur{Yan}}\email{yanshen12@nudt.edu.cn}

\affil[1]{\orgdiv{College of Advanced Interdisciplinary Studies}, \orgname{National University of Defense Technology}, \orgaddress{\city{Changsha}, \postcode{410073}, \state{Hunan}, \country{China}}}

\affil[2]{\orgdiv{Department of Information}, \orgname{Army Medical University}, \orgaddress{\city{Chongqing}, \postcode{400042}, \state{Chongqing}, \country{China}}}

\affil[3]{\orgdiv{College of System Engineering}, \orgname{National University of Defense Technology}, \orgaddress{\city{Changsha}, \postcode{410073}, \state{Hunan}, \country{China}}}

\abstract{Autofocus is necessary for high-throughput and real-time scanning in microscopic imaging. Traditional methods rely on complex hardware or iterative hill-climbing algorithms. Recent learning-based approaches have demonstrated remarkable efficacy in a one-shot setting, avoiding hardware modifications or iterative mechanical lens adjustments. 
However, in this paper, we highlight a significant challenge that the richness of image content can significantly affect autofocus performance. When the image content is sparse, previous autofocus methods, whether traditional climbing-hill or learning-based, tend to fail. 
To tackle this, we propose a content-importance-based solution, named SparseFocus, featuring a novel two-stage pipeline. The first stage measures the importance of regions within the image, while the second stage calculates the defocus distance from selected important regions. To validate our approach and benefit the research community, we collect a large-scale dataset comprising millions of labelled defocused images, encompassing both dense, sparse and extremely sparse scenarios. Experimental results show that SparseFocus surpasses existing methods, effectively handling all levels of content sparsity. Moreover, we integrate SparseFocus into our Whole Slide Imaging (WSI) system that performs well in real-world applications. The code and dataset will be made available upon the publication of this paper.}


\maketitle

\input{secs/introduction}

\input{secs/results}
\input{secs/discussions}

\input{secs/methods}

\input{secs/conclusion}

\bmhead{Data availability}
The datasets captured and analyzed during the current study are available from the corresponding author on request.

\bmhead{Code availability}
The code supporting the findings of this study is available from the corresponding author upon request.

\input{sn-article.bbl}

\bmhead{Acknowledgements}
This study is supported by the Independent Innovation Science Fund of National University of Defense Technology (NO.22-ZZCX-064) and the National Natural Science Foundation of China (No. 62406331).

\bmhead{Author contributions}
Y.-P.Z. conceived the idea, contributed to writing the manuscript, conducted experiments, and developed the WSI system. Q.S. performed experiments, wrote code, analyzed data, and prepared figures. X.-X.F. contributed to code writing, data analysis, and figure drawing. J.H., Y.-K.Z., and Y.-F.Z. were involved in data analysis. C.-F.Z., X.L., and W.-X.W. participated in discussions. D.-D.W. prepared and provided the microscope slide samples. S.Y. contributed to writing, discussions, and data analysis.

\bmhead{Materials \& Correspondence}
Correspondence and requests for materials should be addressed to Yongping Zhai.

\bmhead{Competing interests}
The authors declare no competing interests.

\end{document}

%% file: secs/introduction.tex
\section{Introduction}

Optical microscopes~\cite{balasubramanian2023imagining,davidson2002optical,ma2021comprehensive} are widely used in life sciences~\cite{sijtsema1998confocal,del2022field,ghosh2023viewing}, pathological diagnosis~\cite{chen2011optical,pallen2021advances}, wafer defect detection~\cite{ma2023review}, and other fields~\cite{chen2020microsphere}. 
Traditional optical microscopes typically rely on manual focus, which is inefficient and lacks fast adjustment capabilities. 
In applications requiring rapid focus adjustments, such as long-term live cell imaging~\cite{balasubramanian2023imagining}, the focal plane can shift due to cell movement, necessitating the use of autofocus to maintain cells within the focal plane.
In addition, autofocus is critical in high-throughput applications such as whole slide imaging(WSI) systems ~\cite{li2022comprehensive,zhang2019whole,guo2019openwsi}, which require the rapid acquisition and stitching of thousands of images to achieve composite images with resolutions of billions of pixels.

Over the years, researchers have devised various techniques for autofocus, leading to two dominant methodological streams: active and passive methods.
Active autofocus systems~\cite{zhang2019novel,bian2020autofocusing,han2023novel,jiang2024large,instruments2012perfect,chan2018improving} are characterized by the use of specific optical components to focus on a chosen point or area.
For example, the laser triangulation technique is commonly used for autofocus in photolithography~\cite{zhang2019novel,bian2020autofocusing}, where the angles in a triangle are measured to determine unknown defocus distances.
The Coaxial Defocus Detection method~\cite{han2023novel,jiang2024large} leverages the shape and associated parameters of the laser spot to measure the defocus distance. 
A practical application of this can be seen in the Nikon Perfect Focus System (PFS)~\cite{instruments2012perfect}, which employs near-infrared light to identify a reference plane and subsequently adjust the focus in real time.
Phase detection autofocus (PDAF)~\cite{chan2018improving} mainly involves the design of specific CMOS sensors to calculate the focus plane, as exemplified in Nikon DSLR cameras.
Despite the rapid speed and high robustness offered by active autofocus systems, their considerable cost, large size, and limited precision present significant limitations in numerous applications.

In contrast, passive autofocus methods~\cite{huang2024deformable,chinnasamy2022review,pinkard2019deep} achieve focus by analyzing the captured image, regardless of hardware modifications~\cite{subbarao1995optimal}. They do not require a specific light path or sensor, nor do they emit probing light. More specifically, passive methods can be further classified into two distinct categories: reference-based and non-reference-based.
The most typical example of reference-based methods~\cite{huang2024deformable} is the climbing-hill  algorithm~\cite{chinnasamy2022review}. This method controls the z-axis of the microscope to move up and down to detect contrast changes of the captured image, identifying the high-contrast image that signifies correct focus. Although this method offers high accuracy, its speed is compromised due to the need for multiple mechanical adjustments of the microscope.

Recent advances in microscopy focus on accelerating the autofocus speed to facilitate applications such as dynamic cell tracking~\cite{langehanenberg2009automated} and WSI scanning~\cite{li2022comprehensive}.
This has led to the development of non-reference-based (one-shot) autofocus methods that infer the defocus distance directly from a single captured image using regression schemes, primarily employing neural networks~\cite{pinkard2019deep}.
The pioneering work by \cite{jiang2018transform} introduced a Convolutional Neural Network (CNN) to predict defocus distance, training a ResNet on approximately 130,000 images with varying defocus distances to establish a mapping between captured images and their corresponding defocused distances.
Building on this work, \cite{pinkard2019deep} integrated additional off-axis illumination sources and utilized a Fully Connected Fourier Neural Network (FCFNN) for defocus distance estimation.
Meanwhile, \cite{dastidar2020whole} employed the MobileNetV2 network, comparing pixel-wise intensity differences between two defocused images to predict defocus distance.
Further innovations include \cite{liao2022deep} and \cite{xin2021low}, who both utilized the MobileNetV3 network. \cite{liao2022deep} achieved defocus estimation without requiring additional light sources, while~\cite{xin2021low} incorporated a programmable LED array as the illumination source.
Recently, \cite{li2022learning} introduced a two-step process, involving a defocus classification network to determine the direction of defocus and a subsequent refocusing network to estimate the defocus distance.
Lastly, \cite{gu2023single} proposed the Kernel Distillation Autofocus (KDAF) method, leveraging virtual refocusing to estimate defocus distance.

\begin{figure}[t]
    \centering
    \includegraphics[width=\linewidth]{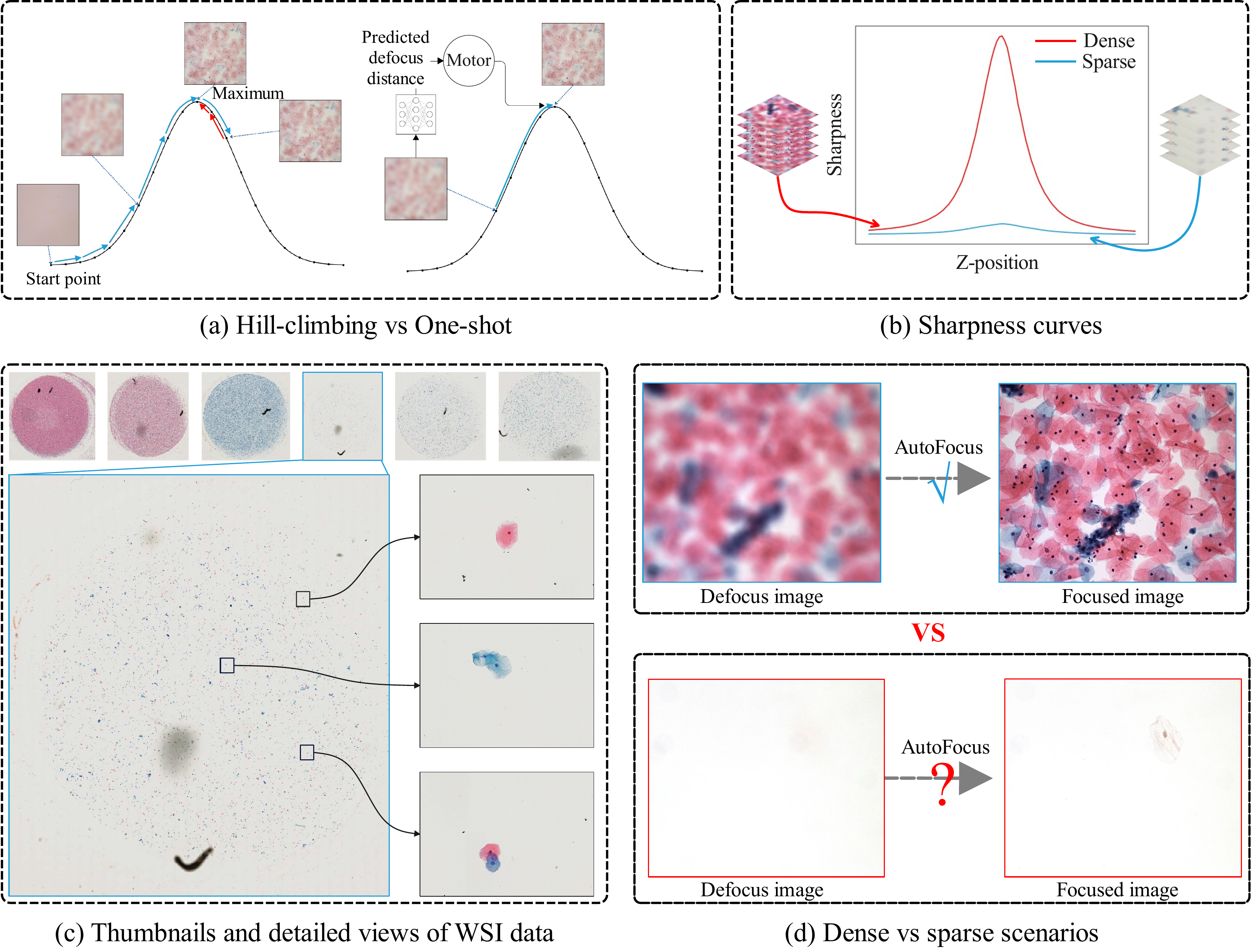}
    \caption{\textbf{The sparsity challenge in microscopy autofocus.} 
    (a) One-shot methods employ neural networks to predict focus distance directly, thus bypassing the iterative adjustments required by traditional hill-climbing techniques.
    (b) illustrates the sharpness curves for both dense and sparse content scenarios. It is evident that the sharpness curve exhibits significant variations in dense content, whereas the changes are less pronounced in sparse content.
    (c) presents images of typical pathological slides, with the upper section showing WSI thumbnails. We display both the cases of dense content and sparse content therein.The first three images are dense content, while the latter three are sparse. To facilitate observation, we enlarge one of the sparse samples and randomly select three representative fields of view, each containing only a few cells.
    (d) This observation raises a critical question: can autofocus be effectively achieved in sparse-content scenarios? }
    \label{fig:teaser}
\end{figure}

In this paper, we highlight a significant challenge that has hardly been noticed in previous methods. This challenge emerges from our observation that in real-world microscopic imaging, the captured content is often sparse, as illustrated in Figure \ref{fig:teaser}. This leads to a majority context of the image being blank, offering no assistance in estimating the defocus distance. Although previous methods employ a patch-based strategy, inferring the focal distance by sampling the observation image into patches and using a voting scheme to achieve a result, our experiments indicate a substantial decline in performance when this situation arises.

To address this, we propose a content-importance-based defocus distance estimation method, named SparseFocus, which is capable of handling autofocus issues across all levels of content sparsity, whether dense, sparse, or extremely sparse. 
This approach overcomes the limitations of current passive focusing methods in dealing with sparse issues, opening up new possibilities for autofocus in practical applicability.
Specifically, we first assign varying importance scores to different regions of the image using a fully convolutional network. Regions with dense content are assigned higher importance scores, regions with sparse content are given lower importance scores, and regions without content are assigned nearly zero importance scores.
After that, we select the top-k regions with the highest importance scores and input them into the defocus distance regression network. 
Based on the defocus distance for each region, we apply a pooling operation to derive the final result.

To train and infer our algorithm, we develop an automated microscopic imaging platform to automatically gather labeled defocused images from microscopic samples. The observational content of this dataset includes pathological tissues and cells, and it accounts for both sparse and dense scenarios. Extensive experiments on this dataset indicate that our method achieves state-of-the-art performance, surpassing previous methods by a significant margin, especially in sparse instances. Notably, even in extremely sparse scenarios, where only one region contains useful content, our method yields satisfactory results.

\bmhead{Contributions}

\begin{itemize}
\item We observe that the richness of image content significantly influences the performance of defocus distance prediction in microscopy. When the image content is sparse, previous autofocus methods, whether traditional climbing-hill or learning-based, tend to fail.
\item We propose a content-importance-based solution featuring a novel two-stage pipeline. The first stage measures the importance of regions within the image, while the second stage calculates the defocus distance from selected important regions.
\item We collect a large-scale dataset comprising millions of defocused images to validate our approach. Experiments indicate that our method significantly outperforms others in both dense and sparse scenarios, and it can also perform autofocus with extremely sparse content.
\item We develop a WSI system equipped with our learning-based one-shot autofocus algorithm, which demonstrates promising focusing capability in real-world scenarios.
\end{itemize}

%% file: secs/results.tex
\section{Results}\label{sec:method}

\subsection{Dataset}\label{dataset}

\bmhead{Data Capturing}

Considering the scarcity of large-scale, labeled dataset suitable for our network training, we have developed an automated system for microscopic image acquisition coupled with an auto-labeling framework.
Please refer to Section~\ref{sec:wsi} for more information about our automated microscopy system.

To improve the sample diversity, we obtain a large collection of pathological tissues and cellular specimens, including samples from the lung, liver, prostate, and exfoliated cervical cells. The cells are stained using Papanicolaou stain, while the tissues are stained with Hematoxylin and Eosin (H\&E).
Specifically, we first collect $400$ samples from patients of different ages and diseases, and then retain $75$ cell samples and $63$ pathological tissue samples, considering the distinct similarity of most samples. After that, we categorize the data collection into $6$ distinct groups according to the type (cell or tissue) and sparsity (dense, sparse or extremely sparse) of the data. Selected samples are illustrated in the supplementary material.
For each sample, we randomly selected 100 fields of view, each with a resolution of $2016 \times 2016$ pixels. To collect defocus data, a sequence of z-stack images was captured at $101$ different defocus distances, with a step size of $0.5\si{\micro\meter}$ ranging from $-25\si{\micro\meter}$ to $+25\si{\micro\meter}$. In total, we obtain $1,324,211$ pathological microscopic images with defocus distance labels.

\bmhead{Statistics}
We categorize the data based on sparsity and type into six groups. For cells, we have $3345$ dense, $2054$ sparse and $1824$ extremely sparse z-stacks. For tissues, we have $2923$ dense, $1442$ sparse and $1723$ extremely sparse z-stacks.
These are then divided into training, validation, and test sets in an 8:1:1 ratio, ensuring a comprehensive and balanced evaluation of our method.

\subsection{Evaluation Protocols}\label{preliminary}

In learning-based autofocusing methods, \textit{Mean Absolute Error (MAE)} of the predicted defocus distance is often employed to evaluate their performance. However, relying solely on the \textit{MAE} for assessment has some limitations. For example, it does not indicate whether the estimated defocus distance falls within the Depth of Field (DoF), nor does it reveal if the defocus direction is incorrectly estimated. As a result, we introduce two additional metrics: \textit{DoF-Accuracy} and \textit{Direction Success Score (DSS)}. The \textit{DoF-Accuracy} evaluates the degree to which the predicted focus point falls within the Depth of Field (DoF), providing a comprehensive measure of how well an algorithm can achieve autofocus, especially for microsystems with varying parameters and distinct DoFs. The \textit{DSS} quantifies the precision of the direction prediction. We introduce this metrics because an incorrect direction prediction by the autofocus algorithm can lead to more severe image defocus, potentially placing the defocus distance outside the system's operational range and preventing successful focus resolution.

The \textit{MAE} quantifies the accuracy of the autofocus algorithm by determining the average absolute difference between the predicted and true defocus distances, which is calculated as:
\begin{equation}
MAE = \frac{1}{|D|} \sum_D ||e_{d_i}|| = \frac{1}{|D|} \sum_D ||\widehat{d_i} - d^*_i||,
\end{equation}
where $e_{d_i}$ represents the absolute error, $\widehat{d_i}$ is the predicted defocus distance, $d^*_i$ is the ground truth defocus distance, and $|D|$ is the number of samples in the test dataset. 

\textit{DoF-Accuracy} measures the percentage of absolute errors $e_{d_i}$ that fall within $1/n$ DoF (in this study, we set $n = \{1, 2, 3\}$), allowing for a fair comparison between optical microscope systems with different DoF. \textit{DoF-Accuracy} is defined as:
\begin{equation}
DoF\text{-}Accuracy = \frac{1}{|D|} \sum_D \mathbb{I}\left(||e_{d_i}|| \le \frac{1}{n} DoF\right) \times 100\%.
\end{equation}
 
The indicator function $\mathbb{I}(\cdot)$ is defined as:
\begin{equation}
\mathbb{I}(P) = \begin{cases} 1 & \text{if } P \text{ is true} \\ 0 & \text{if } P \text{ is false} \end{cases}
\end{equation}

The \textit{DSS} measures the percentage of cases where the predicted direction of defocus aligns with the true direction, which is defined as:

\begin{equation}
DSS = \frac{1}{|D|} \sum_D \mathbb{I}\left(\operatorname{sgn}\left(\widehat{d_i} \cdot d^*_i\right) \ge 0\right) \times 100\%
\end{equation}

where $\operatorname{sgn}(\cdot)$ is the sign function, defined as:

\begin{equation}
\operatorname{sgn}(x) =
\begin{cases}
1 & \text{if } x > 0 \\
0 & \text{if } x = 0 \\
-1 & \text{if } x < 0
\end{cases}
\end{equation}

For the ``boundary'' case where $d^*_i = 0$, indicating that the image is already in focus, the predicted defocus direction, whether positive or negative, can be considered correct.

\subsection{Experimental Results}
We conduct an extensive evaluation of our algorithm's performance utilizing the large-scale dataset that we have assembled.
To demonstrate the performance of our method, we compared it with several learning-based baselines, including those proposed by Dastidar et al.~\cite{dastidar2020whole}, Liao et al.~\cite{liao2022deep}, Jiang et al.~\cite{jiang2018transform} and Li et al.~\cite{li2022learning}.
For each baseline, we partition the image into non-overlapping patches in grid mode, infer the defocusing distance of each patch, and get the result by a median filter operation.
To maintain the fairness of the experiments, all baseline models are trained under the same condition as our proposed method, including the use of the same datasets and hyperparameter settings (such as learning rate, batch size, and number of iterations).

\begin{figure}[htbp]
	\centering
	\includegraphics[width=0.95\linewidth]{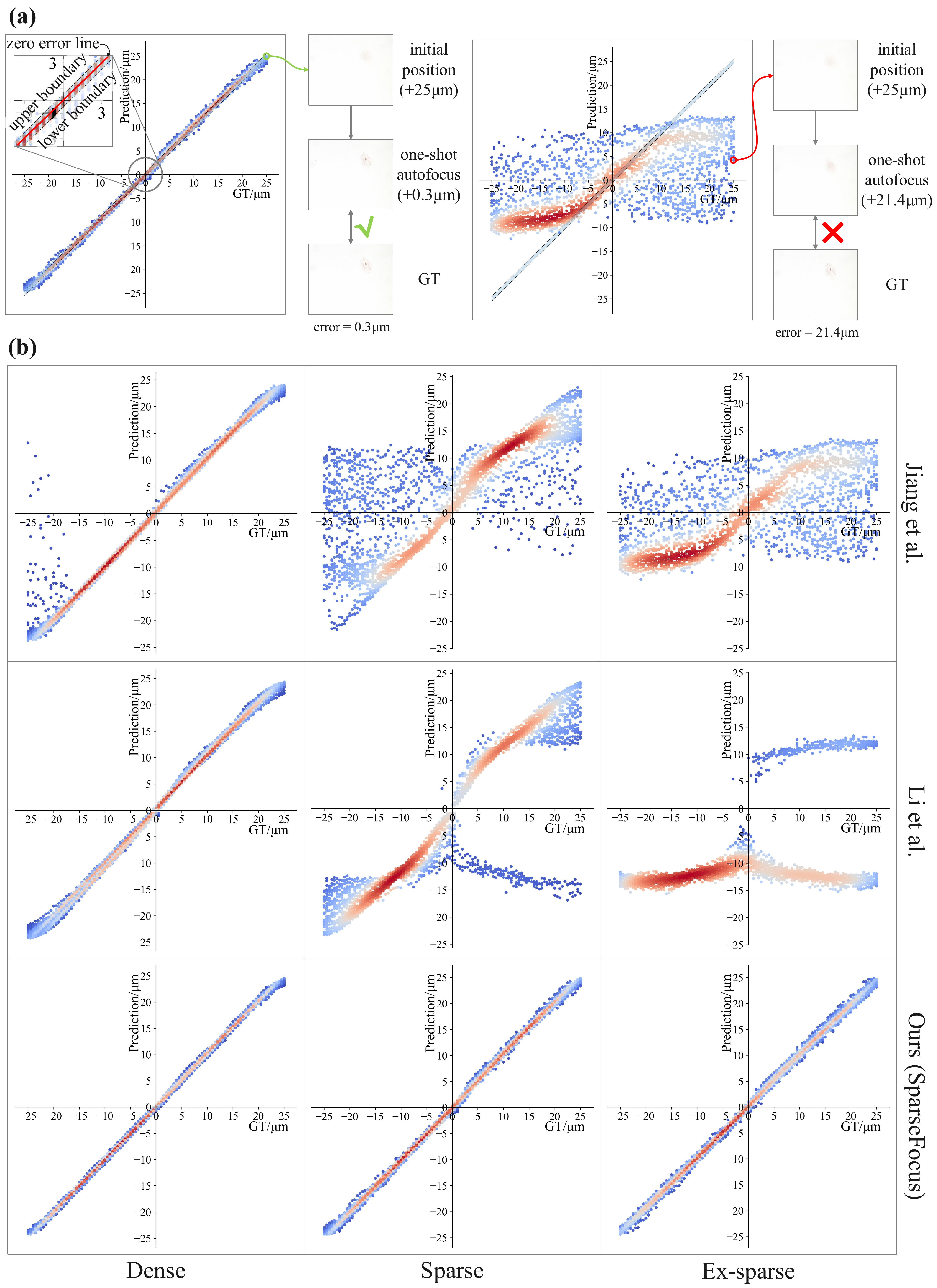}
 	\caption{\textbf{Comparing with baselines using regression plots.} The regression plots show the results, with the horizontal axis representing the ground-truth values and the vertical axis indicating the predicted results. In (a), we mark the upper and lower boundaries of the DoF with lines to illustrate the distribution of results within this range. An ideal zero-error line is included for reference, indicating that the closer a point is to this line, the smaller the error and the better the outcome. Points within the depth of field represent accurate predictions, whereas those outside indicate significant errors. Incorrect predictions are located in the opposite quadrant. In (b), we compare with baselines using such regression plots with different content sparsity.}
	\label{fig:scatters}
\end{figure}

First, we evaluated our method using randomly selected samples from each z-stack, as presented in regression plots in Figure~\ref{fig:scatters}. 
In Figure~\ref{fig:scatters}(a), we provide case study of our method with a baseline method (Jiang et al.~\cite{jiang2018transform}) under extremely sparse conditions, involving only one cell. At a defocus distance of $+25\si{\micro\meter}$, the cell is nearly invisible. Despite this, our method accurately predicts the defocus distance and direction, achieving an error of just $0.3\si{\micro\meter}$.
Figure~\ref{fig:scatters}(b) presents more testing results, encompassing scenarios of high density, sparsity, and extreme sparsity. The test results demonstrate that our method significantly outperforms the baseline, particularly under sparse and extremely sparse conditions. The majority of our method's predictions fall within the DoF, unlike comparing baselines, which exhibit substantial incorrectness and frequent directional errors, rendering them incapable of accurate focusing under such conditions.
More comparative experiments with other baselines are provided in the supplementary materials.

Next, we assess the performance of the proposed method using the three metrics defined in this paper: \textit{MAE (Mean Absolute Error)}, \textit{DoF-Accuracy} and \textit{DSS (Direction Success Score)}. Table~\ref{tab:dde} presents a comparative analysis of the \textit{MAE} for predicted defocus distances. The results demonstrate that our method consistently achieves the lowest \textit{MAE} and variance across all conditions, regardless of the sparsity of the data content.
Notably, for images with sparse content, our method surpasses previous methods by a very large margin, reducing the error rate by an order of magnitude.
This improvement is particularly striking when considering images with extreme sparsity, where our method successfully performs autofocus with remarkable accuracies of $0.60\si{\micro\meter}$ for cells and $0.45\si{\micro\meter}$ for tissues, while other baselines basically fail.
This finding highlights the accuracy, stability, and robustness of the proposed method.

\begin{table}[htbp]
\caption{\textbf{Comparing with baselines using \textit{MAE}.} We report the mean absolute error (\textit{MAE}) and variance across varying levels of sparsity (dense, sparse, extremely sparse).}
\label{tab:dde}
\begin{tabular}{lcccccc}
\toprule
\multirow{2}{*}{MAE ($\si{\micro\meter}\downarrow$)} & \multicolumn{3}{c}{Cell} & \multicolumn{3}{c}{Tissue} \\
\cmidrule(l){2-4} \cmidrule(l){5-7}
 & Dense & Sparse & Ex-sparse & Dense & Sparse & Ex-sparse \\ 
\midrule
Dastidar et al. \cite{dastidar2020whole} & $0.40\pm0.50$ & $4.63\pm5.67$ & $9.79\pm7.68$ & $0.93\pm1.70$ & $6.53\pm6.24$ & $10.05\pm7.41$ \\ 
Liao et al. \cite{liao2022deep} & $0.66\pm1.19$ & $4.93\pm5.79$ & $9.95\pm7.16$ & $1.70\pm5.09$ & $5.94\pm7.12$ & $10.42\pm6.94$ \\ 
Jiang et al. \cite{jiang2018transform} & $0.73\pm1.93$ & $6.19\pm7.90$ & $8.92\pm7.34$ & $1.42\pm4.55$ & $4.88\pm6.62$ & $8.73\pm6.69$ \\ 
Li et al. \cite{li2022learning} & $0.72\pm0.53$ & $3.93\pm6.28$ & $12.25\pm10.65$ & $1.61\pm5.05$ & $6.47\pm7.26$ & $14.24\pm11.32$ \\ 
\textbf{Ours} & $\mathbf{0.37\pm0.29}$ & $\mathbf{0.51\pm0.40}$ & $\mathbf{0.60\pm0.46}$ & $\mathbf{0.49\pm0.40}$ & $\mathbf{0.43\pm0.33}$ & $\mathbf{0.45\pm0.36}$ \\ 
\bottomrule
\end{tabular}
\end{table}

The effectiveness of our method is also substantiated through the \textit{DoF-Accuracy}, as illustrated in Table~\ref{tab:dof_cell}. While achieving comparable performance in dense scenarios, our method exhibits a significant advantage in sparse and extremely sparse cases.
In sparse situations, the defocus distances predicted by other baselines often fall outside the DoF, leading to unsuccessful autofocus.
In contrast, our method consistently achieves accurate focus. For instance, in the extremely sparse case, 58.23\% of cell dataset and 71.71\% of tissue dataset fell within the DoF.

\begin{table}[htbp]
\caption{\textbf{Comparing with baselines using \textit{DoF-Accuracy} for cell (top) and tissue (bottom) samples.} We report the \textit{DoF-Accuracy} at $n = \{1, 2, 3\}$ across different sparsity levels (dense, sparse, extremely sparse).}
\label{tab:dof_cell}
\begin{tabular}{lccccccccc}
\toprule
\multirow{2}{*}{DoF-Accuracy (\% $\uparrow$)} & \multicolumn{3}{c}{Dense} & \multicolumn{3}{c}{Sparse} & \multicolumn{3}{c}{Ex-sparse} \\
\cmidrule(l){2-4} \cmidrule(l){5-7} \cmidrule(l){8-10}
& 1/3 & 1/2 & 1 & 1/3 & 1/2 & 1 & 1/3 & 1/2 & 1 \\
\midrule
\multicolumn{10}{c}{Cell} \\
\midrule
Dastidar et al. \cite{dastidar2020whole} & 35.18 & 50.24 & \textbf{80.95} & 11.93 & 17.20 & 31.49 & 4.07 & 5.75 & 11.27 \\
Liao et al. \cite{liao2022deep} & 23.40 & 33.35 & 62.13 & 10.75 & 15.53 & 28.52 & 2.26 & 3.21 & 7.08 \\
Jiang et al. \cite{jiang2018transform} & 30.95 & 42.54 & 67.15 & 7.90 & 11.34 & 20.64 & 4.61 & 6.34 & 10.95 \\
Li et al. \cite{li2022learning} & 27.46 & 35.10 & 50.12 & 7.41 & 10.07 & 14.39 & 1.85 & 2.76 & 4.20 \\
\textbf{Ours} & \textbf{35.99} & \textbf{50.81} & 80.46 & \textbf{26.58} & \textbf{37.60} & \textbf{66.55} & \textbf{23.18} & \textbf{33.73} & \textbf{58.23} \\
\midrule
\multicolumn{10}{c}{Tissue} \\
\midrule
Dastidar et al. \cite{dastidar2020whole} & 20.48 & 29.83 & 53.37 & 4.28 & 6.75 & 14.50 & 1.75 & 2.88 & 6.19 \\
Liao et al. \cite{liao2022deep} & 20.21 & 29.29 & 55.82 & 7.99 & 12.13 & 25.37 & 1.75 & 2.81 & 4.81 \\
Jiang et al. \cite{jiang2018transform} & 20.81 & 29.14 & 56.35 & 6.87 & 10.33 & 22.41 & 4.06 & 5.19 & 8.63 \\
Li et al. \cite{li2022learning} & 24.83 & 33.03 & 47.42 & 6.08 & 7.66 & 11.04 & 1.81 & 2.44 & 3.75 \\
\textbf{Ours} & \textbf{26.54} & \textbf{39.38} & \textbf{69.72} & \textbf{27.53} & \textbf{40.85} & \textbf{74.69} & \textbf{28.16} & \textbf{41.74} & \textbf{71.71} \\
\bottomrule
\end{tabular}
\end{table}

Table~\ref{tab:da} illustrates the \textit{Direction Success Score (DSS)}. Our method achieves exceptional performance, exceeding 99\% for all conditions. In contrast, other methods exhibit numerous incorrect predictions, particularly in sparse and extremely sparse scenarios.

\begin{table}[htbp]
\caption{\textbf{Comparing with baselines using \textit{DSS}.} We report the direction success score (\textit{DSS}) across different sparsity levels (dense, sparse, extremely sparse).}
\label{tab:da}
\begin{tabular}{lcccccc}
\toprule
\multirow{2}{*}{DSS (\% $\uparrow$)} & \multicolumn{3}{c}{Cell} & \multicolumn{3}{c}{Tissue} \\
\cmidrule(r){2-4}
\cmidrule(l){5-7} & Dense & Sparse & Ex-sparse & Dense & Sparse & Ex-sparse \\
\midrule
Dastidar et al. \cite{dastidar2020whole} & \textbf{100.00} & 95.43 & 69.48 & \textbf{99.95} & 96.13 & 63.38 \\
Liao et al. \cite{liao2022deep} & \textbf{100.00} & 95.64 & 75.92 & 95.49 & 84.38 & 68.15 \\
Jiang et al. \cite{jiang2018transform} & 99.69 & 83.99 & 76.83 & 97.27 & 92.12 & 83.98 \\
Li et al. \cite{li2022learning} & \textbf{100.00}  & 92.89 & 63.23 & 97.46 & 88.86 & 53.41 \\
\textbf{Ours} & \textbf{100.00} & \textbf{99.80} & \textbf{99.96} & 99.77 & \textbf{99.91} & \textbf{99.94}\\ 
\botrule
\end{tabular}
\end{table}

\begin{figure}[htbp]
	\centering
	\includegraphics[width=\linewidth]{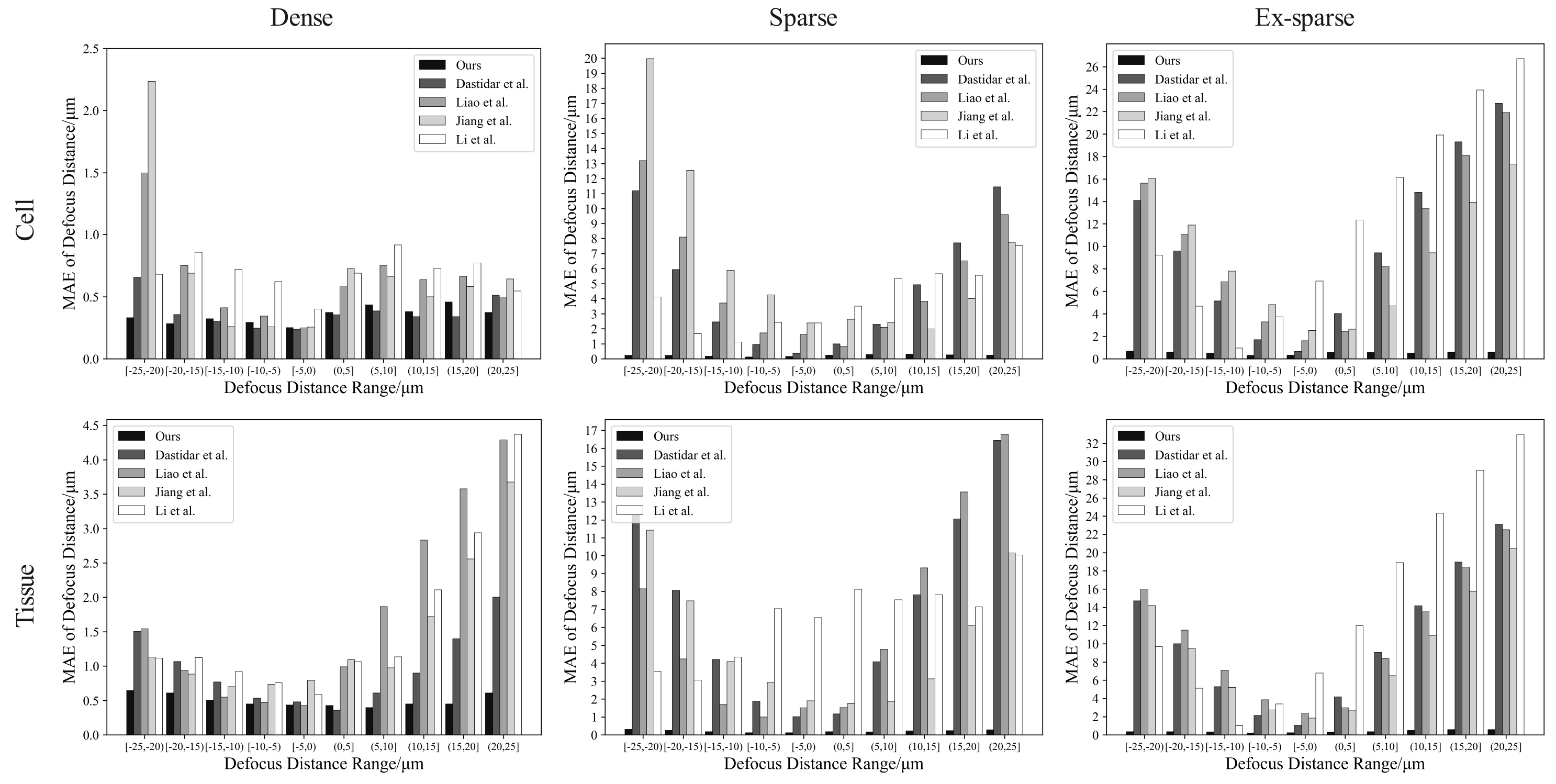}
	\caption{\textbf{Comparing with baselines using \textit{MAE} histogram.} The histogram illustrates the \textit{MAE} performance of our method compared to baseline approaches across various defocus distance for both cell and tissue samples.}
	\label{fig:bars}
\end{figure}

To further evaluate our method's performance across varying defocus distances, we divided the defocus distance into non-overlapping intervals and calculated the \textit{MAE} for each interval. 
Figure~\ref{fig:bars} displays the error distribution histogram. 
It can be observed that the while \textit{MAE} of almost all methods increases with greater defocus distance, the degree of increase varies significantly. In dense scenarios, the \textit{MAE} remains relatively stable across defocus distances. However, in sparse and extremely sparse scenarios, baseline methods exhibit a significant rise in \textit{MAE} as defocus distance increases. 
For example, in sparse tissue samples, the \textit{MAE} of Dastidar et.al~\cite{dastidar2020whole} increases from approximately $\num{1}~\si{\nano\meter}$ to about $\num{16}~\si{\nano\meter}$, resulting in autofocus failure. 
In contrast, our method shows minimal sensitivity to varying defocus distance, consistently maintaining low error levels regardless of the initialization to the focal plane.

%% file: secs/discussions.tex
\section{Discussions}

\subsection{The Asymmetry of the Point Spread Function (PSF) in Microscopy}

\begin{figure}[H]
	\centering
	\includegraphics[width=\linewidth]{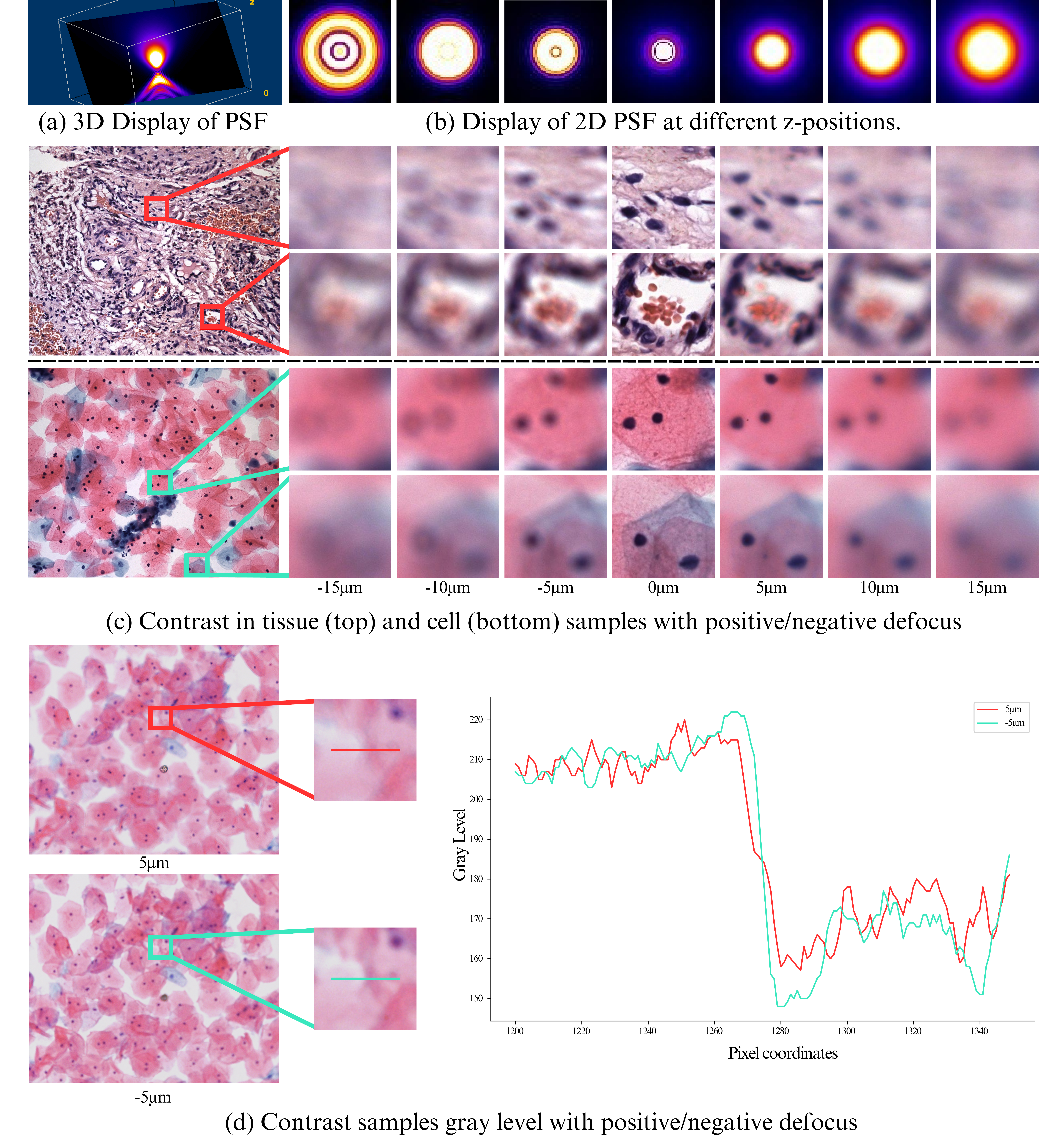}
	\caption{\textbf{The asymmetry of PSF.} (a) and (b) illustrate the PSF of a microscopy imaging system, highlighting its asymmetry with respect to the focal plane. (c) demonstrates the defocused imaging relative to the focal plane, and (d) presents the comparison of the gray values of pixels at the same positions corresponding to 5\si{\micro\meter} and -5\si{\micro\meter}. Both of them provide corroborative evidence for the disparities in images at the corresponding locations.}
	\label{fig:psf}
\end{figure}

In the ideal imaging model, the Point Spread Function (PSF) of a microscope is symmetric with respect to the focal plane. 
This symmetry allows algorithms to estimate the absolute defocus distance but prevents them from determining whether the defocus is above or below the focal plane, thereby rendering one-shot autofocusing seemingly impractical.
However, in real optical microscopy systems, the PSF often exhibits significant asymmetry due to refractive index mismatches among the different media in the imaging path, such as the slide, sample, cover slip, and the surrounding medium like air or immersion oil. 
These mismatches introduce aberrations, including spherical aberration, coma, astigmatism, field curvature, and distortion, which disrupt the ideal symmetric distribution of the PSF. 
Figures~\ref{fig:psf}(a) and \ref{fig:psf}(b) present the 3D PSF and 2D PSF of our developed Whole Slide Imaging (WSI) device.
These visualizations are generated using the Gibson \& Lanni PSF model~\cite{Gibson:89} within the open-source software Fiji~\cite{Schindelin2012-jh} and the PSF Generator plugin\footnote{http://bigwww.epfl.ch/publications/kirshner1103.html}.
Figure~\ref{fig:psf}(c) shows images at symmetric defocus distances on both sides of the focal plane.
Figure~\ref{fig:psf}(d) illustrates the differences in pixel grayscale values at the same position for images at 5\si{\micro\meter} and -5\si{\micro\meter}.
These visualizations also demonstrate that, in real optical microscopy systems, the PSF is asymmetric.
Additional theoretical analysis on the PSF is provided in the supplementary materials.

The asymmetry of the PSF, though potentially detrimental to image quality, presents a unique opportunity for one-shot autofocusing. This phenomenon results in images with positive or negative defocus on either side of the focal plane exhibiting distinct characteristics. Although these differences are subtle, the sophisticated feature extraction capabilities of deep learning can effectively discern them. By capitalizing on this physical principle, we propose a one-shot learning-based network designed to estimate both the defocus distance and direction from a single image.

\subsection{Autofocus for Thick Specimens}

Autofocus is generally designed for a specific focal plane, assuming that most samples exhibit little variation in elevation over a field of view.
However, for very thick samples, such as those resulting from the slicing of pathological sections, different regions within the same field of view may lie on different focal planes (see supplementary material). This can lead to a scenario where focusing on one region causes others to appear blurry, complicating autofocus efforts.
To address such challenges, strategies may contain: 1) Designate a specific region of interest for the autofocus algorithm to target exclusively; 2) Employ z-stack image fusion strategy, capturing and fusing images at various z-axis positions to achieve a uniformly sharp image across the entire field of view.

%% file: secs/methods.tex
\section{Materials and Methods}\label{sec:experiments}

\begin{figure}[htbp]
	\centering
	\includegraphics[width=\linewidth]{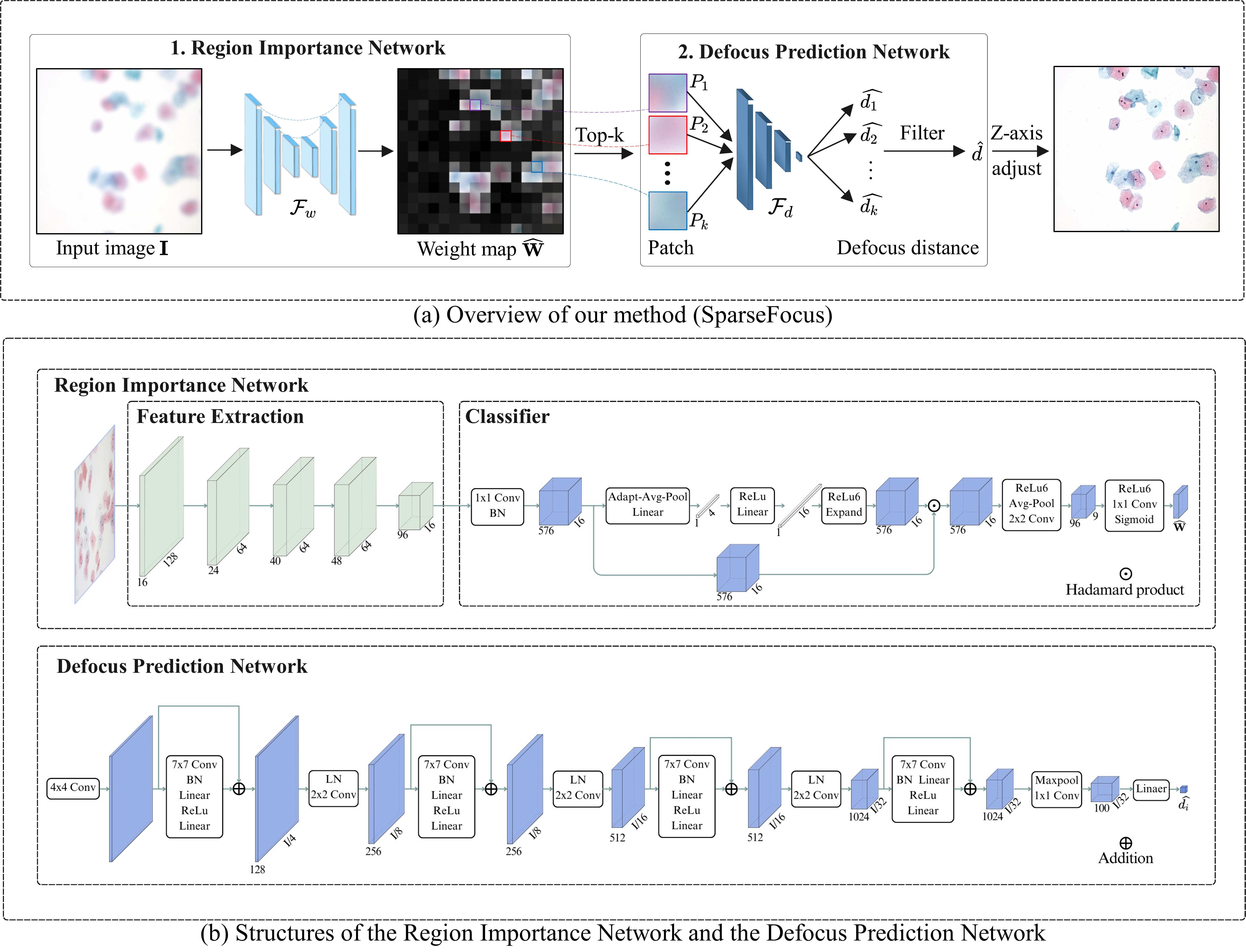}
	\caption{\textbf{Framework of the SparseFocus Network.} (a) is the overview of our method (SparseFocus), (b) shows the structures of the Region Importance Network (RIN) and the Defocus Prediction Network (DPN).}
	\label{fig:framework}
\end{figure}

An overview of our method is shown in Figure \ref{fig:framework}(a). Given a defocused image $\mathbf{I}$ as input, our object is to calculate the defocus distance $\widehat{d}$ for it.
To achieve this goal, we propose a novel two-stage pipeline. In the first stage, the captured defocused image is fed into a fully convolutional network termed Region Importance Network (RIN) $\mathcal{F}_w$, which simultaneously predicts the important weights $\widehat{\mathbf{W}} = \{ \widehat{W}_1, \widehat{W}_2, ..., \widehat{W}_n \}$ for all uniformly-split patches $\mathbf{P} = \{ P_1, P_2, ..., P_n \}$ in the image (detailed in Section~\ref{subsec:network-weight}). In the second stage, Top-k patches $\mathbf{P}_k = \{ P_i \}$ with the highest weights are selected, and a neural network named Defocus Prediction Network (DPN) $\mathcal{F}_d$ is used to predict the defocus distance $\{ \widehat{d}_i = \mathcal{F}_d(P_i) \}$ for each patch. Finally, an aggregate operation, i.e., median filtering, is used to obtain the final result $\widehat{d}$ (detailed in Section~\ref{subsec:network-defocus}). This estimated result $\widehat{d}$ drives the control system, precisely positioning the objective lens at the focus plane for optimal image sharpness.

\subsection{Region Importance Network}\label{subsec:network-weight}

We construct the Region Importance Network (RIN, shown in Figure \ref{fig:framework}(b)) $\mathcal{F}_w$ based on MobileNetV3~\cite{howard2019searching}, which is tailored to assign importance scores to image patches $\mathbf{P} = \{ P_1, P_2, ..., P_n \}$ within a defocused input $\mathbf{I}$.
Specifically, the process begins with cropping the original $2448 \times 2048$ pixel image $\mathbf{I}$ to a $2016 \times 2016$ square, followed by resizing to $512 \times 512$ pixels. This preprocessed image is then fed into RIN, which outputs a $9 \times 9$ prediction matrix $\widehat{\mathbf{W}} = \{ \widehat{W}_1, \widehat{W}_2, ..., \widehat{W}_n \}$. Each element $\widehat{W}_i$ in $\widehat{\mathbf{W}}$ quantifies the content importance of the corresponding $224 \times 224$ pixel patch $P_i$ within the original image $\mathbf{I}$.
We rank all patches $\mathbf{P}$ based on important scores and select the top-$k$ most important patches $\{P_1, P_2,..., P_k\}$ for subsequent processing. The value of $k$ is denoted as $k_d$ for dense, $k_s$ for sparse and $k_{es}$ for extremely sparse scenarios.

\subsection{Defocus Prediction Network}\label{subsec:network-defocus}

We introduce the Defocus Prediction Network (DPN, shown in Figure \ref{fig:framework}(b)) $\mathcal{F}_d$ to predict defocus distance for each selected patch ($\{ P_1, P_2, ..., P_k \}$). Initially, input patches are processed by a $4 \times 4$ convolution with stride 4 and BatchNorm2d.
These features are then refined through multiple DFENet blocks and downsampling modules. In contrast to prior networks that rely on $3 \times 3$ convolutions, the DFENet block utilizes large kernel convolutions ($7 \times 7$), enabling the model to capture comprehensive spatial information and long-range dependencies within the input patches.
Furthermore, a max pooling layer is incorporated to retain the most salient features before they are fed into a linear regression layer for defocus distance prediction $\{\widehat{d}_1, \widehat{d}_2,..., \widehat{d}_k\}$.
At last, a median filtering is employed to return the final result $\widehat{d}$.

\subsection{Supervision and Loss Function}\label{subsec:supervison}

The loss $\mathcal{L}$ contains the losses for Region Importance Network $\mathcal{L}_w$ and Defocus Prediction Network $\mathcal{L}_d$.

\bmhead{Importance Prediction Supervision}
The loss function for the $\mathcal{L}_w$ is the Binary Cross-Entropy (BCE) Loss~\cite{long2015fully}, calculated between the prediction matrix $\widehat{\mathbf{W}}$ and the Content Richness matrix $\mathbf{W}^*$. 
Let $N$ denotes the total number of elements in $\widehat{\mathbf{W}}$ or $\mathbf{W}^*$, and let $i$ and $j$ denote the indices along the $x$ and $y$ axes, respectively, the loss function for $\mathcal{L}_w$ is given by:

\begin{equation}
\mathcal{L}_w = -\frac{1}{N} \sum_{(i,j) \in \mathbf{W}^*} \left[ \mathbf{W}^*_{i,j} \log{\widehat{\mathbf{W}}_{i,j}} + (1 - \mathbf{W}^*_{i,j}) \log{(1 - \widehat{\mathbf{W}}_{i,j})} \right]
\end{equation}

\bmhead{Defocus Prediction Supervision} 
The loss function for $\mathcal{L}_d$ is the $\mathcal
{L}_2$ loss. For each selected patch $P_i \in \{ P_1, P_2, ..., P_k \}$, we obtain the true defocus distances $d^*_i \in \{ d^*_1, d^*_2, ..., d^*_k \}$. For true distance labeling, please refer to supplementary materials for more details. The loss function for $\mathcal{L}_d$ is:

\begin{equation}
\mathcal{L}_d = \frac{1}{M} \sum_{D} (\widehat{d}_i - d^*_i)^2
\end{equation}

where $M$ is the number of patches labeled as positive in the predicted matrix $\widehat{\mathbf{W}}$. $D = {\widehat{d}_1, \widehat{d}_2, \ldots, \widehat{d}_k}$ represents the predicted values from the Defocus Prediction Network.

\subsection{Implement Details}\label{subsec:implement}

During training, we augmented the data diversity by adjusting image brightness, contrast, and saturation. Brightness was varied within the range [0.9, 1.4], while contrast and saturation were adjusted within [0.8, 1.5]. Our model was trained for 600 epochs on four NVIDIA RTX 3090 GPUs. We used the Adam optimizer with an initial learning rate of $5 \times 10^{-5}$ and a batch size of 16 for the Region Importance Network. For the Defocus Prediction Network, we employed an initial learning rate of $1 \times 10^{-4}$ and a batch size of 128. We set the importance threshold $\rho$ to 0.8, and the number of selected patches $k_d$, $k_s$ and $k_{es}$ are set to $k_d=31$, $k_d=9$ and $k_{es}=3$.

\section{One-shot Autofocus enhanced WSI system}
\label{sec:wsi}

In this section, we introduce an advanced one-shot autofocus-enhanced Whole Slide Imaging system (osa-WSI), based on our learning-based autofocus algorithm, coupled with an efficient image stitching protocol for large-scale imaging and an online motion path planning. The osa-WSI encompasses several key components: 

\begin{itemize}
\item Motorized XY and Z Stages: These stages offer a resolution of 100 nm in the XY plane and 50 nm in the Z-axis, with repeatability of 400 nm in XY and 150 nm in Z, ensuring precise focusing within the microscope's depth of field.
\item Macro-Imaging System: This subsystem captures slide thumbnails and identifies specimen regions,  using both transmissive and reflective illumination modes. 
\item Microscopy Imaging System: This subsystem is designed for high-fidelity imaging of both transparent and opaque samples, equipped with a LED white light source, an APO objective lens, and a global shutter CMOS camera.
\item Motion Control System: This subsystem excels in path planning and camera triggering, enhancing the overall imaging workflow.
\item Integrated Algorithms: These include autofocus, image stitching, AI-based recognition, and image quality assessment.
\end{itemize}

\begin{figure}[H]
	\centering
	\includegraphics[width=\linewidth]{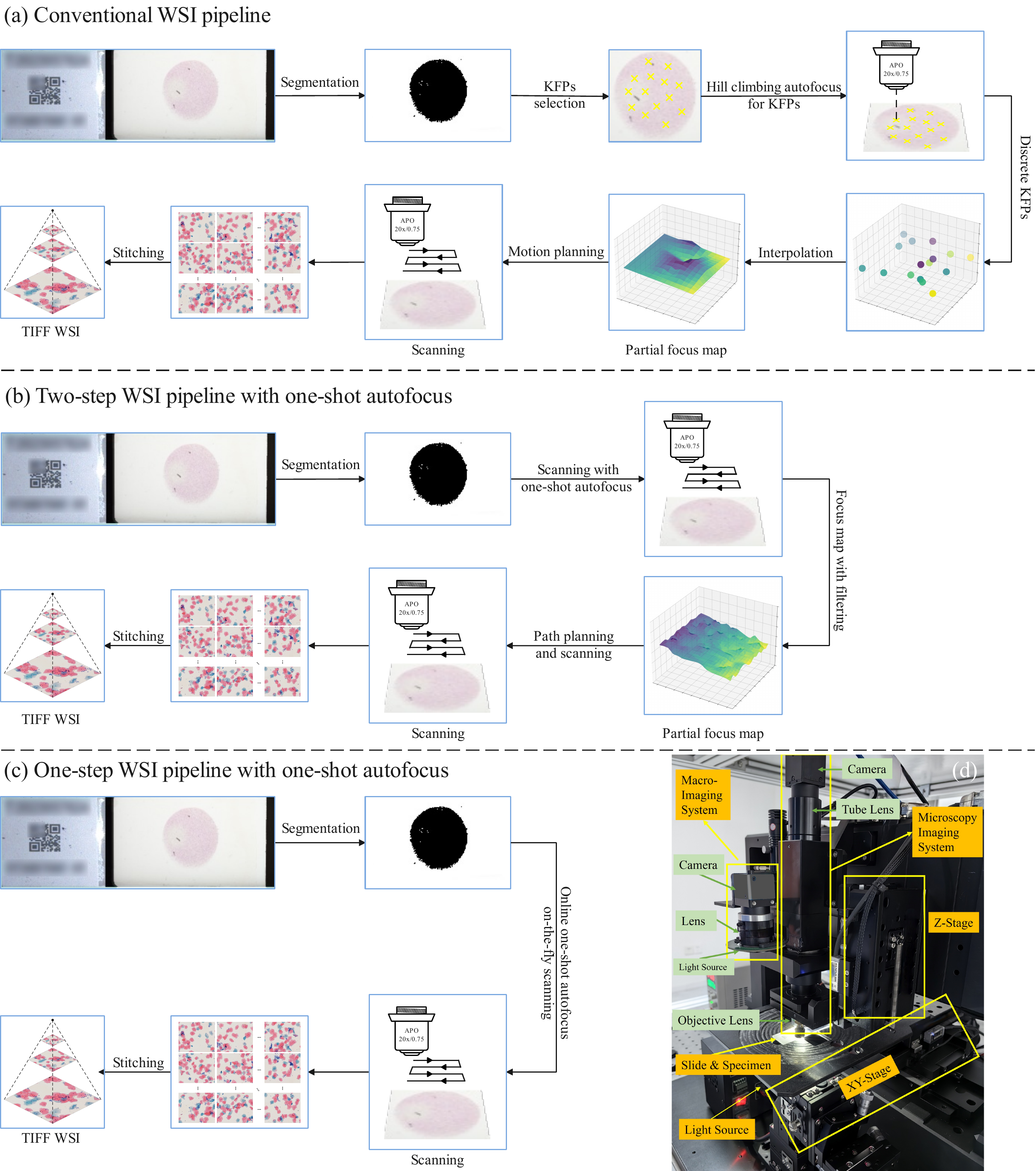}
	\caption{\textbf{Overview of three WSI pipelines.} (a) represents the workflow of a traditional WSI system. (b) illustrates the pipeline incorporating one-shot autofocus technology, implemented via a two-step method. (c) demonstrates the dynamic acquisition process of the WSI pipeline using one-shot autofocus. (d) showcases the real instrumentation of our one-shot autofocus enhanced WSI system.
    }
	\label{fig:net}
\end{figure}

\begin{figure}[H]
	\centering
	\includegraphics[width=\linewidth]{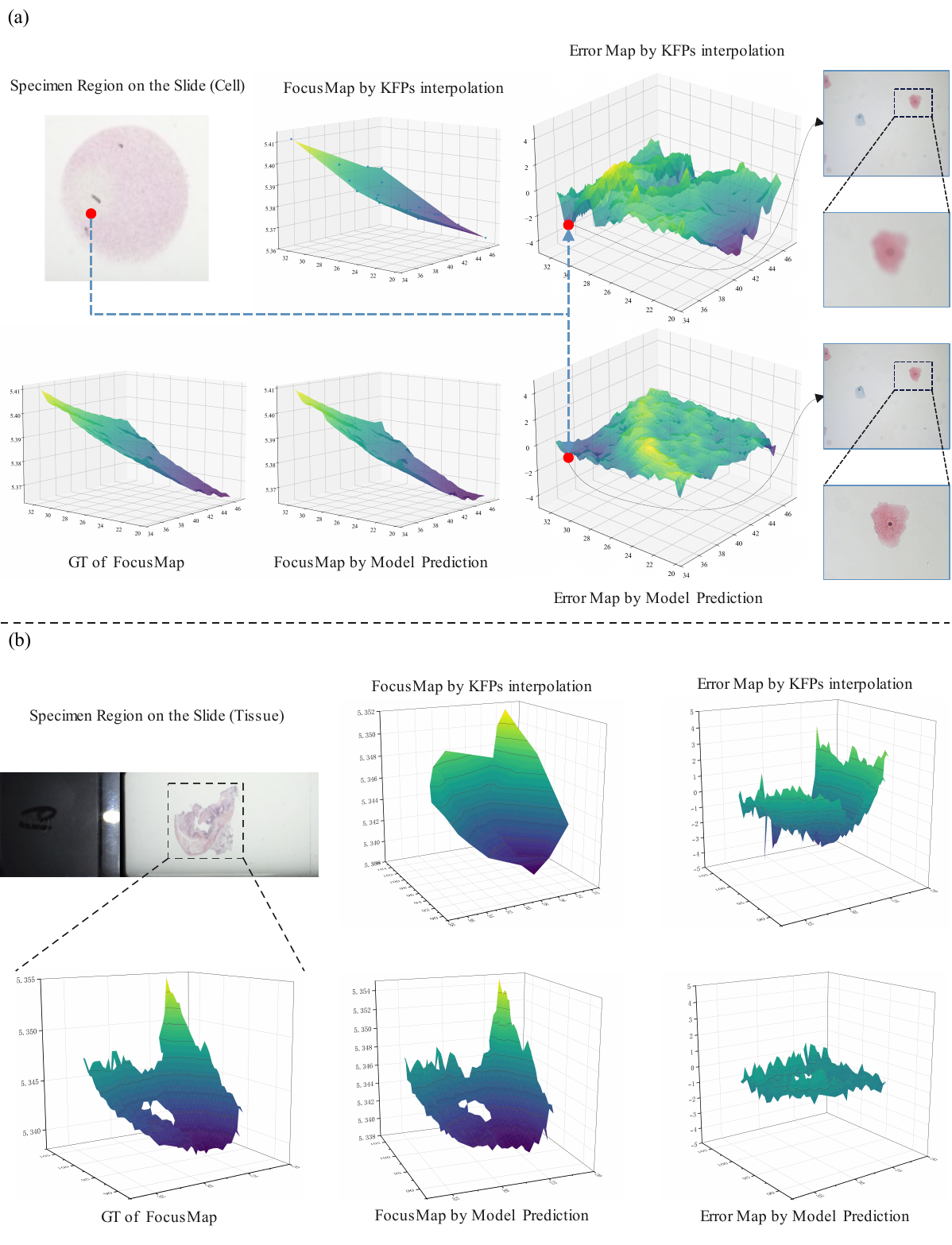}
    \caption{\textbf{Autofocus across the entire slice using our osa-WSI system.}  We present two cases—cells in panel (a) and tissue in panel (b)—to demonstrate the superior autofocus capability of our osa-WSI system. Readers can compare the accuracy of our method with the traditional KFP interpolation approach by examining the error maps provided in the figures.}
	\label{fig:wsi}
\end{figure}

These technologies establish our WSI system as a cutting-edge tool in the field of pathology.

Figure~\ref{fig:net} presents a comparison of the workflow pipelines between the proposed osa-WSI and a traditional WSI system. The traditional system, depicted in Figure~\ref{fig:net}(a), relies on interpolating from pre-selected Key Focus Points (KFPs) to create a FocusMap for the entire slide, which is inefficient and susceptible to significant errors on uneven sample surfaces. 
Our osa-WSI system addresses these issues, as outlined in Figures~\ref{fig:net}(b) and \ref{fig:net}(c). Figure~\ref{fig:net}(b) describes a two-step process that maintains the FocusMap construction but with improved speed and accuracy. In contrast, Figure~\ref{fig:net}(c) illustrates a one-step process that generates FocusMaps in real-time during scanning, thereby eliminating the need for pre-construction.

Furthermore, Figure~\ref{fig:wsi} provides two examples of defocus distance estimation across the entire slice using the proposed osa-WSI, termed the FocusMap. This method outperforms the traditional KFP interpolation technique in focal estimation accuracy, as evidenced by the error map comparisons.

%% file: secs/conclusion.tex
\section{Conclusion}
\label{sec:conclusion}

We propose a learning-based two-stage network incorporating Region Importance Network and Defocus Prediction Network to enable autofocus for both dense, sparse or even extremely sparse microscopy images under the one-shot setting.
Our method can effectively identify local regions with important content and subsequently estimate the defocus distance using these regions.
We also introduce a new large-scale dataset that provides a variety of defocused images with dense and sparse scenarios.
Experimental results demonstrate that our method significantly improves defocus distance estimation accuracy compared to existing learning-based one-shot methods, especially in sparse scenarios.
Building on this method, we develop a WSI system that exhibits promising focusing capabilities in real-world applications.

%% file: sn-article.bbl

%% file: sn-article.bbl
\begin{thebibliography}{34}
\ifx \bisbn   \undefined \def \bisbn  #1{ISBN #1}\fi
\ifx \binits  \undefined \def \binits#1{#1}\fi
\ifx \bauthor  \undefined \def \bauthor#1{#1}\fi
\ifx \batitle  \undefined \def \batitle#1{#1}\fi
\ifx \bjtitle  \undefined \def \bjtitle#1{#1}\fi
\ifx \bvolume  \undefined \def \bvolume#1{\textbf{#1}}\fi
\ifx \byear  \undefined \def \byear#1{#1}\fi
\ifx \bissue  \undefined \def \bissue#1{#1}\fi
\ifx \bfpage  \undefined \def \bfpage#1{#1}\fi
\ifx \blpage  \undefined \def \blpage #1{#1}\fi
\ifx \burl  \undefined \def \burl#1{\textsf{#1}}\fi
\ifx \doiurl  \undefined \def \doiurl#1{\url{https://doi.org/#1}}\fi
\ifx \betal  \undefined \def \betal{\textit{et al.}}\fi
\ifx \binstitute  \undefined \def \binstitute#1{#1}\fi
\ifx \binstitutionaled  \undefined \def \binstitutionaled#1{#1}\fi
\ifx \bctitle  \undefined \def \bctitle#1{#1}\fi
\ifx \beditor  \undefined \def \beditor#1{#1}\fi
\ifx \bpublisher  \undefined \def \bpublisher#1{#1}\fi
\ifx \bbtitle  \undefined \def \bbtitle#1{#1}\fi
\ifx \bedition  \undefined \def \bedition#1{#1}\fi
\ifx \bseriesno  \undefined \def \bseriesno#1{#1}\fi
\ifx \blocation  \undefined \def \blocation#1{#1}\fi
\ifx \bsertitle  \undefined \def \bsertitle#1{#1}\fi
\ifx \bsnm \undefined \def \bsnm#1{#1}\fi
\ifx \bsuffix \undefined \def \bsuffix#1{#1}\fi
\ifx \bparticle \undefined \def \bparticle#1{#1}\fi
\ifx \barticle \undefined \def \barticle#1{#1}\fi
\bibcommenthead
\ifx \bconfdate \undefined \def \bconfdate #1{#1}\fi
\ifx \botherref \undefined \def \botherref #1{#1}\fi
\ifx \url \undefined \def \url#1{\textsf{#1}}\fi
\ifx \bchapter \undefined \def \bchapter#1{#1}\fi
\ifx \bbook \undefined \def \bbook#1{#1}\fi
\ifx \bcomment \undefined \def \bcomment#1{#1}\fi
\ifx \oauthor \undefined \def \oauthor#1{#1}\fi
\ifx \citeauthoryear \undefined \def \citeauthoryear#1{#1}\fi
\ifx \endbibitem  \undefined \def \endbibitem {}\fi
\ifx \bconflocation  \undefined \def \bconflocation#1{#1}\fi
\ifx \arxivurl  \undefined \def \arxivurl#1{\textsf{#1}}\fi
\csname PreBibitemsHook\endcsname

\bibitem[\protect\citeauthoryear{Balasubramanian et~al.}{2023}]{balasubramanian2023imagining}
\begin{barticle}
\bauthor{\bsnm{Balasubramanian}, \binits{H.}},
\bauthor{\bsnm{Hobson}, \binits{C.M.}},
\bauthor{\bsnm{Chew}, \binits{T.-L.}},
\bauthor{\bsnm{Aaron}, \binits{J.S.}}:
\batitle{Imagining the future of optical microscopy: everything, everywhere, all at once}.
\bjtitle{Communications Biology}
\bvolume{6}(\bissue{1}),
\bfpage{1096}
(\byear{2023})
\end{barticle}
\endbibitem

\bibitem[\protect\citeauthoryear{Davidson and Abramowitz}{2002}]{davidson2002optical}
\begin{barticle}
\bauthor{\bsnm{Davidson}, \binits{M.W.}},
\bauthor{\bsnm{Abramowitz}, \binits{M.}}:
\batitle{Optical microscopy}.
\bjtitle{Encyclopedia of imaging science and technology}
\bvolume{2}(\bissue{1106-1141}),
\bfpage{120}
(\byear{2002})
\end{barticle}
\endbibitem

\bibitem[\protect\citeauthoryear{Ma and Fei}{2021}]{ma2021comprehensive}
\begin{barticle}
\bauthor{\bsnm{Ma}, \binits{L.}},
\bauthor{\bsnm{Fei}, \binits{B.}}:
\batitle{Comprehensive review of surgical microscopes: technology development and medical applications}.
\bjtitle{Journal of biomedical optics}
\bvolume{26}(\bissue{1}),
\bfpage{010901}--\blpage{010901}
(\byear{2021})
\end{barticle}
\endbibitem

\bibitem[\protect\citeauthoryear{Sijtsema et~al.}{1998}]{sijtsema1998confocal}
\begin{barticle}
\bauthor{\bsnm{Sijtsema}, \binits{N.M.}},
\bauthor{\bsnm{Wouters}, \binits{S.D.}},
\bauthor{\bsnm{Grauw}, \binits{C.J.D.}},
\bauthor{\bsnm{Otto}, \binits{C.}},
\bauthor{\bsnm{Greve}, \binits{J.}}:
\batitle{Confocal direct imaging raman microscope: design and applications in biology}.
\bjtitle{Applied spectroscopy}
\bvolume{52}(\bissue{3}),
\bfpage{348}--\blpage{355}
(\byear{1998})
\end{barticle}
\endbibitem

\bibitem[\protect\citeauthoryear{Del~Rosario et~al.}{2022}]{del2022field}
\begin{barticle}
\bauthor{\bsnm{Del~Rosario}, \binits{M.}},
\bauthor{\bsnm{Heil}, \binits{H.S.}},
\bauthor{\bsnm{Mendes}, \binits{A.}},
\bauthor{\bsnm{Saggiomo}, \binits{V.}},
\bauthor{\bsnm{Henriques}, \binits{R.}}:
\batitle{The field guide to 3d printing in optical microscopy for life sciences}.
\bjtitle{Advanced Biology}
\bvolume{6}(\bissue{4}),
\bfpage{2100994}
(\byear{2022})
\end{barticle}
\endbibitem

\bibitem[\protect\citeauthoryear{Ghosh and Agarwal}{2023}]{ghosh2023viewing}
\begin{barticle}
\bauthor{\bsnm{Ghosh}, \binits{B.}},
\bauthor{\bsnm{Agarwal}, \binits{K.}}:
\batitle{Viewing life without labels under optical microscopes}.
\bjtitle{Communications Biology}
\bvolume{6}(\bissue{1}),
\bfpage{559}
(\byear{2023})
\end{barticle}
\endbibitem

\bibitem[\protect\citeauthoryear{Chen et~al.}{2011}]{chen2011optical}
\begin{barticle}
\bauthor{\bsnm{Chen}, \binits{X.}},
\bauthor{\bsnm{Zheng}, \binits{B.}},
\bauthor{\bsnm{Liu}, \binits{H.}}:
\batitle{Optical and digital microscopic imaging techniques and applications in pathology}.
\bjtitle{Analytical Cellular Pathology}
\bvolume{34}(\bissue{1-2}),
\bfpage{5}--\blpage{18}
(\byear{2011})
\end{barticle}
\endbibitem

\bibitem[\protect\citeauthoryear{Pallen et~al.}{2021}]{pallen2021advances}
\begin{barticle}
\bauthor{\bsnm{Pallen}, \binits{S.}},
\bauthor{\bsnm{Shetty}, \binits{Y.}},
\bauthor{\bsnm{Das}, \binits{S.}},
\bauthor{\bsnm{Vaz}, \binits{J.M.}},
\bauthor{\bsnm{Mazumder}, \binits{N.}}:
\batitle{Advances in nonlinear optical microscopy techniques for in vivo and in vitro neuroimaging}.
\bjtitle{Biophysical Reviews}
\bvolume{13}(\bissue{6}),
\bfpage{1199}--\blpage{1217}
(\byear{2021})
\end{barticle}
\endbibitem

\bibitem[\protect\citeauthoryear{Ma et~al.}{2023}]{ma2023review}
\begin{barticle}
\bauthor{\bsnm{Ma}, \binits{J.}},
\bauthor{\bsnm{Zhang}, \binits{T.}},
\bauthor{\bsnm{Yang}, \binits{C.}},
\bauthor{\bsnm{Cao}, \binits{Y.}},
\bauthor{\bsnm{Xie}, \binits{L.}},
\bauthor{\bsnm{Tian}, \binits{H.}},
\bauthor{\bsnm{Li}, \binits{X.}}:
\batitle{Review of wafer surface defect detection methods}.
\bjtitle{Electronics}
\bvolume{12}(\bissue{8}),
\bfpage{1787}
(\byear{2023})
\end{barticle}
\endbibitem

\bibitem[\protect\citeauthoryear{Chen et~al.}{2020}]{chen2020microsphere}
\begin{botherref}
\oauthor{\bsnm{Chen}, \binits{L.}},
\oauthor{\bsnm{Zhou}, \binits{Y.}},
\oauthor{\bsnm{Zhou}, \binits{R.}},
\oauthor{\bsnm{Hong}, \binits{M.}}:
Microsphere—toward future of optical microscopes.
Iscience
\textbf{23}(6)
(2020)
\end{botherref}
\endbibitem

\bibitem[\protect\citeauthoryear{Li et~al.}{2022}]{li2022comprehensive}
\begin{barticle}
\bauthor{\bsnm{Li}, \binits{X.}},
\bauthor{\bsnm{Li}, \binits{C.}},
\bauthor{\bsnm{Rahaman}, \binits{M.M.}},
\bauthor{\bsnm{Sun}, \binits{H.}},
\bauthor{\bsnm{Li}, \binits{X.}},
\bauthor{\bsnm{Wu}, \binits{J.}},
\bauthor{\bsnm{Yao}, \binits{Y.}},
\bauthor{\bsnm{Grzegorzek}, \binits{M.}}:
\batitle{A comprehensive review of computer-aided whole-slide image analysis: from datasets to feature extraction, segmentation, classification and detection approaches}.
\bjtitle{Artificial Intelligence Review}
\bvolume{55}(\bissue{6}),
\bfpage{4809}--\blpage{4878}
(\byear{2022})
\end{barticle}
\endbibitem

\bibitem[\protect\citeauthoryear{Zhang et~al.}{2019}]{zhang2019whole}
\begin{bchapter}
\bauthor{\bsnm{Zhang}, \binits{Y.}},
\bauthor{\bsnm{Chen}, \binits{H.}},
\bauthor{\bsnm{Wei}, \binits{Y.}},
\bauthor{\bsnm{Zhao}, \binits{P.}},
\bauthor{\bsnm{Cao}, \binits{J.}},
\bauthor{\bsnm{Fan}, \binits{X.}},
\bauthor{\bsnm{Lou}, \binits{X.}},
\bauthor{\bsnm{Liu}, \binits{H.}},
\bauthor{\bsnm{Hou}, \binits{J.}},
\bauthor{\bsnm{Han}, \binits{X.}}, \betal:
\bctitle{From whole slide imaging to microscopy: Deep microscopy adaptation network for histopathology cancer image classification}.
In: \bbtitle{International Conference on Medical Image Computing and Computer-assisted Intervention},
pp. \bfpage{360}--\blpage{368}
(\byear{2019}).
\bcomment{Springer}
\end{bchapter}
\endbibitem

\bibitem[\protect\citeauthoryear{Guo et~al.}{2019}]{guo2019openwsi}
\begin{barticle}
\bauthor{\bsnm{Guo}, \binits{C.}},
\bauthor{\bsnm{Bian}, \binits{Z.}},
\bauthor{\bsnm{Jiang}, \binits{S.}},
\bauthor{\bsnm{Murphy}, \binits{M.}},
\bauthor{\bsnm{Zhu}, \binits{J.}},
\bauthor{\bsnm{Wang}, \binits{R.}},
\bauthor{\bsnm{Song}, \binits{P.}},
\bauthor{\bsnm{Shao}, \binits{X.}},
\bauthor{\bsnm{Zhang}, \binits{Y.}},
\bauthor{\bsnm{Zheng}, \binits{G.}}:
\batitle{Openwsi: a low-cost, high-throughput whole slide imaging system via single-frame autofocusing and open-source hardware}.
\bjtitle{Optics Letters}
\bvolume{45}(\bissue{1}),
\bfpage{260}--\blpage{263}
(\byear{2019})
\end{barticle}
\endbibitem

\bibitem[\protect\citeauthoryear{Zhang et~al.}{2019}]{zhang2019novel}
\begin{barticle}
\bauthor{\bsnm{Zhang}, \binits{X.}},
\bauthor{\bsnm{Fan}, \binits{F.}},
\bauthor{\bsnm{Gheisari}, \binits{M.}},
\bauthor{\bsnm{Srivastava}, \binits{G.}}:
\batitle{A novel auto-focus method for image processing using laser triangulation}.
\bjtitle{Ieee Access}
\bvolume{7},
\bfpage{64837}--\blpage{64843}
(\byear{2019})
\end{barticle}
\endbibitem

\bibitem[\protect\citeauthoryear{Bian et~al.}{2020}]{bian2020autofocusing}
\begin{barticle}
\bauthor{\bsnm{Bian}, \binits{Z.}},
\bauthor{\bsnm{Guo}, \binits{C.}},
\bauthor{\bsnm{Jiang}, \binits{S.}},
\bauthor{\bsnm{Zhu}, \binits{J.}},
\bauthor{\bsnm{Wang}, \binits{R.}},
\bauthor{\bsnm{Song}, \binits{P.}},
\bauthor{\bsnm{Zhang}, \binits{Z.}},
\bauthor{\bsnm{Hoshino}, \binits{K.}},
\bauthor{\bsnm{Zheng}, \binits{G.}}:
\batitle{Autofocusing technologies for whole slide imaging and automated microscopy}.
\bjtitle{Journal of Biophotonics}
\bvolume{13}(\bissue{12}),
\bfpage{202000227}
(\byear{2020})
\end{barticle}
\endbibitem

\bibitem[\protect\citeauthoryear{Han et~al.}{2023}]{han2023novel}
\begin{barticle}
\bauthor{\bsnm{Han}, \binits{C.}},
\bauthor{\bsnm{Tang}, \binits{Y.}},
\bauthor{\bsnm{Cheng}, \binits{X.}},
\bauthor{\bsnm{Sun}, \binits{H.}},
\bauthor{\bsnm{Feng}, \binits{J.}},
\bauthor{\bsnm{Hu}, \binits{S.}}:
\batitle{A novel coaxial focus position detection technique based on differential modulation evaluation for laser direct photolithography}.
\bjtitle{Optics and Lasers in Engineering}
\bvolume{161},
\bfpage{107396}
(\byear{2023})
\end{barticle}
\endbibitem

\bibitem[\protect\citeauthoryear{Jiang et~al.}{2024}]{jiang2024large}
\begin{bchapter}
\bauthor{\bsnm{Jiang}, \binits{R.}},
\bauthor{\bsnm{Zhang}, \binits{X.}},
\bauthor{\bsnm{Hu}, \binits{X.}},
\bauthor{\bsnm{Dong}, \binits{B.}}:
\bctitle{Large dynamic range of secondary position errors detection for the coaxial three-mirror system using dual-branch convolutional neural network}.
In: \bbtitle{Sixth Conference on Frontiers in Optical Imaging and Technology: Novel Detector Technologies},
vol. \bseriesno{13154},
pp. \bfpage{42}--\blpage{51}
(\byear{2024}).
\bcomment{SPIE}
\end{bchapter}
\endbibitem

\bibitem[\protect\citeauthoryear{Silfies et~al.}{}]{instruments2012perfect}
\begin{botherref}
\oauthor{\bsnm{Silfies}, \binits{J.}},
\oauthor{\bsnm{Lieser}, \binits{E.}},
\oauthor{\bsnm{Stanley}, \binits{S.}},
\oauthor{\bsnm{Davidson}, \binits{M.}}:
{The Nikon Perfect Focus System (PFS)}.
\url{https://www.microscopyu.com/tutorials/the-nikon-perfect-focus-system-pfs}.
Accessed: 2024-11-18
\end{botherref}
\endbibitem

\bibitem[\protect\citeauthoryear{Chan and Chen}{2018}]{chan2018improving}
\begin{barticle}
\bauthor{\bsnm{Chan}, \binits{C.-C.}},
\bauthor{\bsnm{Chen}, \binits{H.H.}}:
\batitle{Improving the reliability of phase detection autofocus}.
\bjtitle{Electronic Imaging}
\bvolume{30},
\bfpage{1}--\blpage{5}
(\byear{2018})
\end{barticle}
\endbibitem

\bibitem[\protect\citeauthoryear{Huang et~al.}{2024}]{huang2024deformable}
\begin{botherref}
\oauthor{\bsnm{Huang}, \binits{H.}},
\oauthor{\bsnm{Liu}, \binits{Y.}},
\oauthor{\bsnm{Siewerdsen}, \binits{J.H.}},
\oauthor{\bsnm{Lu}, \binits{A.}},
\oauthor{\bsnm{Hu}, \binits{Y.}},
\oauthor{\bsnm{Zbijewski}, \binits{W.}},
\oauthor{\bsnm{Unberath}, \binits{M.}},
\oauthor{\bsnm{Weiss}, \binits{C.R.}},
\oauthor{\bsnm{Sisniega}, \binits{A.}}:
Deformable motion compensation in interventional cone-beam ct with a context-aware learned autofocus metric.
Medical physics
(2024)
\end{botherref}
\endbibitem

\bibitem[\protect\citeauthoryear{Chinnasamy et~al.}{2022}]{chinnasamy2022review}
\begin{botherref}
\oauthor{\bsnm{Chinnasamy}, \binits{S.}},
\oauthor{\bsnm{Ramachandran}, \binits{M.}},
\oauthor{\bsnm{Amudha}, \binits{M.}},
\oauthor{\bsnm{Ramu}, \binits{K.}}:
A review on hill climbing optimization methodology.
Recent Trends in Management and Commerce
\textbf{3}(1)
(2022)
\end{botherref}
\endbibitem

\bibitem[\protect\citeauthoryear{Pinkard et~al.}{2019}]{pinkard2019deep}
\begin{barticle}
\bauthor{\bsnm{Pinkard}, \binits{H.}},
\bauthor{\bsnm{Phillips}, \binits{Z.}},
\bauthor{\bsnm{Babakhani}, \binits{A.}},
\bauthor{\bsnm{Fletcher}, \binits{D.A.}},
\bauthor{\bsnm{Waller}, \binits{L.}}:
\batitle{Deep learning for single-shot autofocus microscopy}.
\bjtitle{Optica}
\bvolume{6}(\bissue{6}),
\bfpage{794}--\blpage{797}
(\byear{2019})
\end{barticle}
\endbibitem

\bibitem[\protect\citeauthoryear{Subbarao and Tyan}{1995}]{subbarao1995optimal}
\begin{bchapter}
\bauthor{\bsnm{Subbarao}, \binits{M.}},
\bauthor{\bsnm{Tyan}, \binits{J.}}:
\bctitle{Optimal focus measure for passive autofocusing and depth-from-focus}.
In: \bbtitle{Videometrics IV},
vol. \bseriesno{2598},
pp. \bfpage{89}--\blpage{99}
(\byear{1995}).
\bcomment{SPIE}
\end{bchapter}
\endbibitem

\bibitem[\protect\citeauthoryear{Langehanenberg et~al.}{2009}]{langehanenberg2009automated}
\begin{barticle}
\bauthor{\bsnm{Langehanenberg}, \binits{P.}},
\bauthor{\bsnm{Ivanova}, \binits{L.}},
\bauthor{\bsnm{Bernhardt}, \binits{I.}},
\bauthor{\bsnm{Ketelhut}, \binits{S.}},
\bauthor{\bsnm{Vollmer}, \binits{A.}},
\bauthor{\bsnm{Dirksen}, \binits{D.}},
\bauthor{\bsnm{Georgiev}, \binits{G.}},
\bauthor{\bsnm{Bally}, \binits{G.}},
\bauthor{\bsnm{Kemper}, \binits{B.}}:
\batitle{Automated three-dimensional tracking of living cells by digital holographic microscopy}.
\bjtitle{Journal of biomedical optics}
\bvolume{14}(\bissue{1}),
\bfpage{014018}--\blpage{014018}
(\byear{2009})
\end{barticle}
\endbibitem

\bibitem[\protect\citeauthoryear{Jiang et~al.}{2018}]{jiang2018transform}
\begin{barticle}
\bauthor{\bsnm{Jiang}, \binits{S.}},
\bauthor{\bsnm{Liao}, \binits{J.}},
\bauthor{\bsnm{Bian}, \binits{Z.}},
\bauthor{\bsnm{Guo}, \binits{K.}},
\bauthor{\bsnm{Zhang}, \binits{Y.}},
\bauthor{\bsnm{Zheng}, \binits{G.}}:
\batitle{Transform-and multi-domain deep learning for single-frame rapid autofocusing in whole slide imaging}.
\bjtitle{Biomedical optics express}
\bvolume{9}(\bissue{4}),
\bfpage{1601}--\blpage{1612}
(\byear{2018})
\end{barticle}
\endbibitem

\bibitem[\protect\citeauthoryear{Dastidar and Ethirajan}{2020}]{dastidar2020whole}
\begin{barticle}
\bauthor{\bsnm{Dastidar}, \binits{T.R.}},
\bauthor{\bsnm{Ethirajan}, \binits{R.}}:
\batitle{Whole slide imaging system using deep learning-based automated focusing}.
\bjtitle{Biomedical Optics Express}
\bvolume{11}(\bissue{1}),
\bfpage{480}--\blpage{491}
(\byear{2020})
\end{barticle}
\endbibitem

\bibitem[\protect\citeauthoryear{Liao et~al.}{2022}]{liao2022deep}
\begin{barticle}
\bauthor{\bsnm{Liao}, \binits{J.}},
\bauthor{\bsnm{Chen}, \binits{X.}},
\bauthor{\bsnm{Ding}, \binits{G.}},
\bauthor{\bsnm{Dong}, \binits{P.}},
\bauthor{\bsnm{Ye}, \binits{H.}},
\bauthor{\bsnm{Wang}, \binits{H.}},
\bauthor{\bsnm{Zhang}, \binits{Y.}},
\bauthor{\bsnm{Yao}, \binits{J.}}:
\batitle{Deep learning-based single-shot autofocus method for digital microscopy}.
\bjtitle{Biomedical Optics Express}
\bvolume{13}(\bissue{1}),
\bfpage{314}--\blpage{327}
(\byear{2022})
\end{barticle}
\endbibitem

\bibitem[\protect\citeauthoryear{Xin et~al.}{2021}]{xin2021low}
\begin{barticle}
\bauthor{\bsnm{Xin}, \binits{K.}},
\bauthor{\bsnm{Jiang}, \binits{S.}},
\bauthor{\bsnm{Chen}, \binits{X.}},
\bauthor{\bsnm{He}, \binits{Y.}},
\bauthor{\bsnm{Zhang}, \binits{J.}},
\bauthor{\bsnm{Wang}, \binits{H.}},
\bauthor{\bsnm{Liu}, \binits{H.}},
\bauthor{\bsnm{Peng}, \binits{Q.}},
\bauthor{\bsnm{Zhang}, \binits{Y.}},
\bauthor{\bsnm{Ji}, \binits{X.}}:
\batitle{Low-cost whole slide imaging system with single-shot autofocusing based on color-multiplexed illumination and deep learning}.
\bjtitle{Biomedical Optics Express}
\bvolume{12}(\bissue{9}),
\bfpage{5644}--\blpage{5657}
(\byear{2021})
\end{barticle}
\endbibitem

\bibitem[\protect\citeauthoryear{Li et~al.}{2022}]{li2022learning}
\begin{barticle}
\bauthor{\bsnm{Li}, \binits{Q.}},
\bauthor{\bsnm{Liu}, \binits{X.}},
\bauthor{\bsnm{Han}, \binits{K.}},
\bauthor{\bsnm{Guo}, \binits{C.}},
\bauthor{\bsnm{Jiang}, \binits{J.}},
\bauthor{\bsnm{Ji}, \binits{X.}},
\bauthor{\bsnm{Wu}, \binits{X.}}:
\batitle{Learning to autofocus in whole slide imaging via physics-guided deep cascade networks}.
\bjtitle{Optics Express}
\bvolume{30}(\bissue{9}),
\bfpage{14319}--\blpage{14340}
(\byear{2022})
\end{barticle}
\endbibitem

\bibitem[\protect\citeauthoryear{Gu}{2023}]{gu2023single}
\begin{barticle}
\bauthor{\bsnm{Gu}, \binits{Y.}}:
\batitle{Single-shot focus estimation for microscopy imaging with kernel distillation}.
\bjtitle{IEEE Transactions on Computational Imaging}
\bvolume{9},
\bfpage{542}--\blpage{550}
(\byear{2023})
\end{barticle}
\endbibitem

\bibitem[\protect\citeauthoryear{Gibson and Lanni}{1989}]{Gibson:89}
\begin{barticle}
\bauthor{\bsnm{Gibson}, \binits{S.F.}},
\bauthor{\bsnm{Lanni}, \binits{F.}}:
\batitle{Diffraction by a circular aperture as a model for three-dimensional optical microscopy}.
\bjtitle{J. Opt. Soc. Am. A}
\bvolume{6}(\bissue{9}),
\bfpage{1357}--\blpage{1367}
(\byear{1989})
\doiurl{10.1364/JOSAA.6.001357}
\end{barticle}
\endbibitem

\bibitem[\protect\citeauthoryear{Schindelin et~al.}{2012}]{Schindelin2012-jh}
\begin{barticle}
\bauthor{\bsnm{Schindelin}, \binits{J.}},
\bauthor{\bsnm{Arganda-Carreras}, \binits{I.}},
\bauthor{\bsnm{Frise}, \binits{E.}},
\bauthor{\bsnm{Kaynig}, \binits{V.}},
\bauthor{\bsnm{Longair}, \binits{M.}},
\bauthor{\bsnm{Pietzsch}, \binits{T.}},
\bauthor{\bsnm{Preibisch}, \binits{S.}},
\bauthor{\bsnm{Rueden}, \binits{C.}},
\bauthor{\bsnm{Saalfeld}, \binits{S.}},
\bauthor{\bsnm{Schmid}, \binits{B.}},
\bauthor{\bsnm{Tinevez}, \binits{J.-Y.}},
\bauthor{\bsnm{White}, \binits{D.J.}},
\bauthor{\bsnm{Hartenstein}, \binits{V.}},
\bauthor{\bsnm{Eliceiri}, \binits{K.}},
\bauthor{\bsnm{Tomancak}, \binits{P.}},
\bauthor{\bsnm{Cardona}, \binits{A.}}:
\batitle{Fiji: an open-source platform for biological-image analysis}.
\bjtitle{Nature Methods}
\bvolume{9}(\bissue{7}),
\bfpage{676}--\blpage{682}
(\byear{2012})
\end{barticle}
\endbibitem

\bibitem[\protect\citeauthoryear{Howard et~al.}{2019}]{howard2019searching}
\begin{bchapter}
\bauthor{\bsnm{Howard}, \binits{A.}},
\bauthor{\bsnm{Sandler}, \binits{M.}},
\bauthor{\bsnm{Chu}, \binits{G.}},
\bauthor{\bsnm{Chen}, \binits{L.-C.}},
\bauthor{\bsnm{Chen}, \binits{B.}},
\bauthor{\bsnm{Tan}, \binits{M.}},
\bauthor{\bsnm{Wang}, \binits{W.}},
\bauthor{\bsnm{Zhu}, \binits{Y.}},
\bauthor{\bsnm{Pang}, \binits{R.}},
\bauthor{\bsnm{Vasudevan}, \binits{V.}}, \betal:
\bctitle{Searching for mobilenetv3}.
In: \bbtitle{Proceedings of the IEEE/CVF International Conference on Computer Vision},
pp. \bfpage{1314}--\blpage{1324}
(\byear{2019})
\end{bchapter}
\endbibitem

\bibitem[\protect\citeauthoryear{Long et~al.}{2015}]{long2015fully}
\begin{bchapter}
\bauthor{\bsnm{Long}, \binits{J.}},
\bauthor{\bsnm{Shelhamer}, \binits{E.}},
\bauthor{\bsnm{Darrell}, \binits{T.}}:
\bctitle{Fully convolutional networks for semantic segmentation}.
In: \bbtitle{Proceedings of the IEEE Conference on Computer Vision and Pattern Recognition},
pp. \bfpage{3431}--\blpage{3440}
(\byear{2015})
\end{bchapter}
\endbibitem

\end{thebibliography}
